\documentclass[letterpaper]{article} 
\usepackage{aaai25}  
\usepackage{times}  
\usepackage[titletoc]{appendix}
\usepackage{helvet}  
\usepackage{courier}  
\usepackage[hyphens]{url}  
\usepackage{graphicx} 
\usepackage{natbib}  
\usepackage{caption} 
\usepackage{algorithm}
\usepackage{algorithmic}
 
\usepackage{newfloat}
\usepackage{listings}
\DeclareCaptionStyle{ruled}{labelfont=normalfont,labelsep=colon,strut=off} 
\floatstyle{ruled}
\newfloat{listing}{tb}{lst}{}
\floatname{listing}{Listing}
\usepackage{color}
\usepackage{multirow}

\title{A $1000\times$ Faster LLM-enhanced Algorithm For Path Planning \\ in Large-scale Grid Maps}

\author {
    Junlin Zeng,  
    Xin Zhang,  
    Xiang Zhao,  
    Yan Pan\thanks{Corresponding author}
}
\affiliations {
    National University of Defense Technology, \\
    Science and Technology on Information Systems Engineering Laboratory, \\
    Changsha, China \\
    \{zengjunlinnudt, zhangxin, zhaoxiang, panyan\}@nudt.edu.cn
}

\usepackage{bibentry}
\nocopyright
\begin{document}

\maketitle

\begin{abstract}
Path planning in grid maps, arising from various applications, has garnered significant attention. Existing methods, such as A*, Dijkstra, and their variants, work well for small-scale maps but fail to address large-scale ones due to high search time and memory consumption. Recently, Large Language Models (LLMs) have shown remarkable performance in path planning but still suffer from spatial illusion and poor planning performance. Among all the works, LLM-A* \cite{meng2024llm} leverages LLM to generate a series of waypoints and then uses A* to plan the paths between the neighboring waypoints. In this way, the complete path is constructed. However, LLM-A* still suffers from high computational time for large-scale maps. To fill this gap, we conducted a deep investigation into LLM-A* and found its bottleneck, resulting in limited performance. Accordingly, we design an innovative LLM-enhanced algorithm, abbr. as iLLM-A*. iLLM-A* includes 3 carefully designed mechanisms, including the optimization of A*, an incremental learning method for LLM to generate high-quality waypoints, and the selection of the appropriate waypoints for A* for path planning. Finally, a comprehensive evaluation on various grid maps shows that, compared with LLM-A*, iLLM-A* \textbf{1) achieves more than $1000\times$ speedup on average, and up to $2349.5\times$ speedup in the extreme case, 2) saves up to $58.6\%$ of the memory cost, 3) achieves both obviously shorter path length and lower path length standard deviation.}


\end{abstract}

%

\section{Introduction}

Path planning in a grid map determines a collision-free path from a start location to a goal location, adhering to specific criteria such as minimizing distance, time, or energy \cite{liu2023path}. This is a fundamental problem in a wide range of real-world applications, such as robot navigation \cite{carvalho2025deep}, automated vehicle parking \cite {jiang2023grid}, and player role planning in game or emulated training environments \cite{panov2018grid}. 

Existing algorithms such as A*, Dijkstra, and their variants are capable of finding the optimal path with the time complexity of $O(N^4log N)$ \cite{carlson2023optimal}, where $N$ is the edge length (grid number) of a 2D square grid map and $N^2$ is the total grid number of the map. Such algorithms work well for small-scale maps. However, in more scenarios, the need for path planning for large-scale grid maps boosts \cite{sun2024multi}. An example is that, with the enhancement of robots' capacities, their working space dramatically expands \cite{tang2025large}. Another example is that with the proliferation of high-resolution computer games, the game maps are increasingly complex
\cite{lee2013fast,kirilenko2025generative}. In such large-scale grid maps, the existing algorithms encounter a significant computation cost increase in both time and memory \cite{ou2022improved}. In recent years, Large Language Models (LLMs) have achieved a remarkable milestone in addressing various planning tasks, inspired by their notable reasoning and planning capacities in complex contexts. Specifically, several works have utilized LLMs for path planning \cite{fan2025embodied}, which, however, suffer from spatial illusion \cite{aghzal2024can, Xie2024,kwon2024language} and result in unstable and limited planning performance. 

To achieve robust path planning, a state-of-the-art (SOTA) work \cite{meng2024llm} proposed LLM-A*, which combined the global insight of LLM with the robust planning capacity of A*. The basic idea of LLM-A* is that the LLM first generates a series of waypoints between the start and goal locations, then A* iteratively plans the paths between the neighbor waypoints, and finally, the whole path is constructed by connecting all the waypoints in sequence. In this way, A* does not need to explore the entire map when planning a path between each neighboring waypoints. Therefore, the overall computational and memory cost is reduced. However, when being applied for large-scale grid maps (with $N\geq 200$), LLM-A* suffers from some critical limitations. First, the implementation of A* in LLM-A*, such as the grid cost query and collision detection, is inefficient, resulting in a long search time. Second, the global OPEN and CLOSED lists enconter high memory cost. Third, LLMs may stochastically generate some inappropriate waypoints. LLM-A* naively utilizes all waypoints for path planning, while some waypoints may be redundant and could be reduced for a better path.

Motivated by LLM-A*, this work proposes an innovative LLM-enhanced path planning algorithm, abbr. as iLLM-A*. Specifically, iLLM-A* consists of three remarkable mechanisms.
1) \textbf{Optimization of A*}: To reduce the search time of A*, we first use a hash table to replace the linear CLOSED list of A* to store the explored grids for fast grid query, then update the evaluation values of a small portion (instead of all of the unexplored grids) of the OPEN list, and finally use an efficient two-stage collision detection to replace the precise collision detection to reduce the detection cost. 2) \textbf{Waypoint Generating by Incremental Learning based LLM}:  We implement an incremental learning-based prompt, in which the Few-shot prompt is dynamically enriched, to guide the LLM to generate higher quality waypoints.
3) \textbf{Appropriate Waypoint Selection}: To address redundant LLM-generated waypoints, we employ an experience-driven method to select the appropriate subset of waypoints from the LLM-generated waypoints for path planning. 
Finally, a comprehensive evaluation on various grid maps shows that, compared to LLM-A *, iLLM-A* \textbf{1) achieves more than $1000\times$ speedup on average and up to $2349.5\times$ speedup in extreme cases, 2) {saves up to $58.6\%$ of the memory cost}, 3) achieves both obviously shorter path length and lower path length standard deviation.}

\section{Limitation Analysis for LLM-A*}

\begin{figure}[t]
    \centering
    \includegraphics[width=\columnwidth]{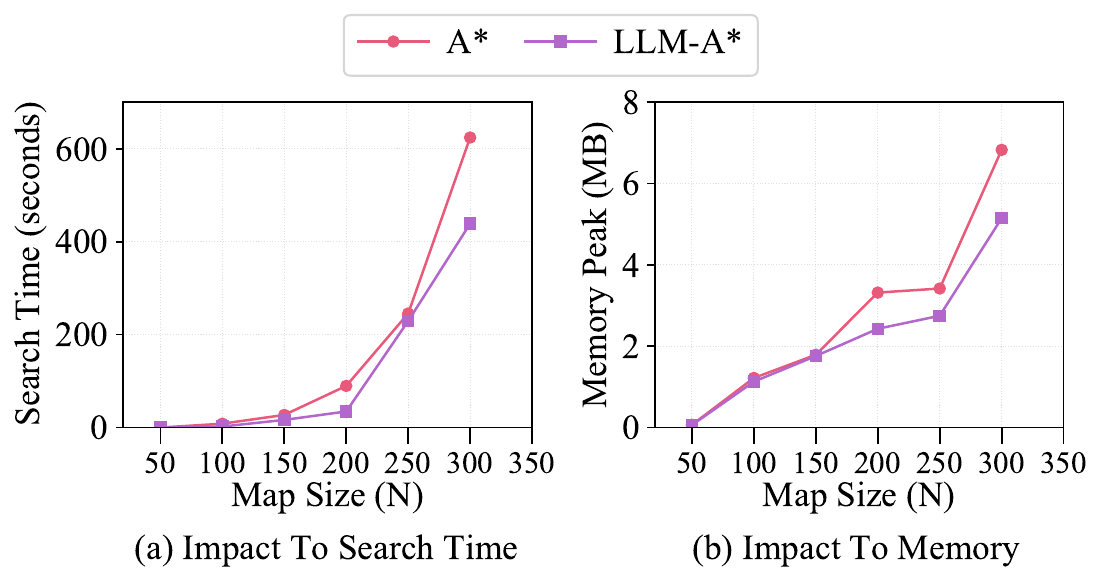}
    \caption{Search Time and Memory of LLM-A* and A* Given Different Map Size.}
    \label{fig:TimeEva}
\end{figure}

We first test the performance of LLM-A* in large-scale grid maps. For simplicity, the work considers a 2D square grid map with equal edges, whose edge length (grid number) is denoted as $N$. The evaluation settings are identical to those in Sec.\ref{sec:eva} and omitted here. Fig. \ref{fig:TimeEva} depicts the search time and memory of A* and LLM-A* given different map sizes, where the memory is the maximum occupied memory of the algorithm. As Fig. \ref{fig:TimeEva} (a) shows, LLM-A* achieves a slightly shorter search time than A*. In addition, when the map scales from $N=200$ to $N=300$, the computation time of LLM-A* exponentially increases from $30s$ to $7min$, which shows the inefficiency of LLM-A* for large-scale maps. Fig. \ref{fig:TimeEva} (b) illustrates that the memory shows a similar increasing trend with search time. When the map scales from $N=50$ to $N=300$, whose area increases by $(300/50)^2=36\times$, the memory increases by $84.75\times$($62.5 KB$ vs. $5.17 MB$). The reasons for the inefficiency of LLM-A* are analyzed as follows.

\subsubsection{Limitation 1: Inefficient Implementation of A*.} The A* in LLM-A* uses two linear data structures for the OPEN and CLOSED lists, respectively. The total time complexity of operating the OPEN list (including inserting and sorting the explored grids into the OPEN list) is $O(N_{wp} N_{open}log N_{open})$, where $N_{wp}$ is the number of waypoints generated by LLM and $N_{open}$ is the length of the OPEN list. The total time complexity of checking whether the grids in OPEN are in CLOSED is $O( N_{open} N_{closed})$. Another key operation on the OPEN list is to detect the collision of the path from the grid with the lowest cost to the next waypoint, whose total complexity is $O( N_{open} N_{closed}N_{ob})$. Specifically, $N_{ob}$ is the obstacle number. In summary, the total time complexity of A* in LLM-A* is $O(N_{wp} N_{open}log N_{open}+N_{open} N_{closed}N_{ob})$. In the worst case, $N_{open}$ and $N_{closed}$ may approach $N^2$. Thereafter, the time complexity of LLM-A* is consistent with that of A*, as verified in Fig.\ref{fig:TimeEva} (a). 

\subsubsection{Limitation 2: Global High Memory Cost.} When planning the path between any two neighboring waypoints, A* needs to maintain a global OPEN list and a CLOSED list, which results in heavy memory cost. Therefore, the memory is only slightly lighter than that of A*, which is verified in Fig.\ref{fig:TimeEva} (b).

\subsubsection{Limitation 3: Inherent Limitation of LLM.} Some waypoints generated by LLM are stochastic and cannot precisely capture the objective of path planning, which is called the space illusion \cite{Huang2023ASO} of LLM. The space illusion may be caused by various factors, such as the training data, the training process, and the reasoning process. In the path planning task, LLM imitates the text syntax instead of exactly learning to find the shortest collision-free path. Therefore, LLM generates some inappropriate waypoints. Valmeekam et al. \cite{valmeekam2023planning} verified the ability of LLM in task planning. The work finds that the ratio of the most advanced LLM successfully planning a mission is only 12\%, which shows the limited capacity of LLM and verifies that LLM may generate some redundant waypoints in the path-planning task.

\section{Design of iLLM-A*}
With the limitations of LLM-A*, this section accordingly proposes an innovative design abbr. as iLLM-A*. We next introduce its three core mechanisms: (I) Optimization of A*, (II) Waypoint Generating by Incremental Learning-Based LLM, and (III) Appropriate Waypoint Selection.
\subsection{Optimization of A*}

\subsubsection{Optimization of The CLOSED List.} We replace the traditional linear CLOSED list structure with a hash-based set to improve efficiency. The hash function directly maps each explored grid to its storage location, and thus significantly improves the search speed of A* ~\cite{sun2009simple}. The hash-based structure reduces the complexity of the search operation on the CLOSED list from $O(N_{closed})$ to $O(1)$ ~\cite{sun2007fringe}. In this way, the search cost of planning the path by A* could be dramatically reduced in large-scale maps.

\subsubsection{Optimization of The OPEN List.} The A* in LLM-A* utilizes a heuristic function to estimate the cost of each grid to the destination: $f(s) = g(s) + h(s) + cost(s, s_{wp})$, where $s$ is the grid, $s_{wp}$ is a waypoint, $g(s)$ is the cost from the start, $h(s)$ is the estimated cost to the goal, and $cost(s,s_{wp})$ is the estimated cost from $s$ to $s_{wp}$. The global OPEN list stores $f(s)$ of all explored grids. When A* in LLM-A* plans a path between two new neighbor grids, $cost(s, s_{wp})$ is changed and $f(s)$ in the OPEN list should be updated. However, the update overhead is huge due to the large size of the global OPEN list. iLLM-A* leverages a similar delayed heuristic update strategy ~\cite{sun2009simple}. Specifically, iLLM-A* only updates the heuristic function in the following two cases. Case 1: only the top-$k$ (i.e., $k=100$) grids with the lowest estimated function in the OPEN list are updated. Case 2: when we extract a grid from the OPEN list, its heuristic function is updated if its function is outdated, and the heuristic function $f(s)$ is estimated using the current value.

\subsubsection{Collision Detection Acceleration.} 
\begin{figure}[t]
    \centering
    \includegraphics[width=\columnwidth]{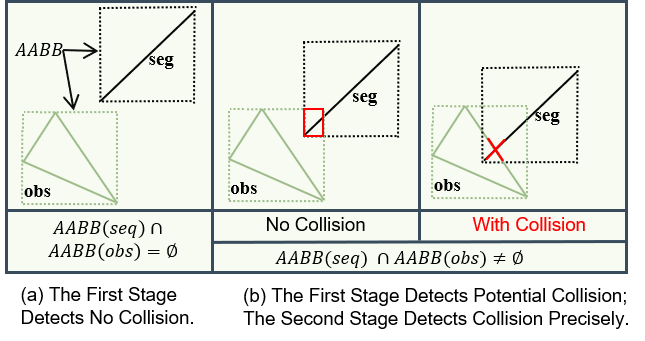}
    \caption{Two-Stage Collision Detection.}
    \label{fig:collision_detection}
\end{figure}

The precise collision detection in LLM-A* is time-consuming. We implement a two-stage collision detection method to reduce the computation overhead. Specifically, we use an Axis-Aligned Bounding Box (AABB) ~\cite{zhu2024developing} to define the minimal rectangle with coordinate-aligned edges completely enclosing a path segment or an obstacle. The AABB edges are parallel to the map edges. The first stage detects potential collisions by testing the overlap of the AABBs of a path segment and an obstacle. Fig.\ref{fig:collision_detection} (a) shows an example in which the two AABBs do not overlap and the path segment would not collide with the obstacle: $AABB(seg)\cap AABB(obs)= \emptyset$. If the two AABBs overlap with each other: $AABB(seg)\cap AABB(obs)\neq \emptyset$, a potential collision is detected, as the two examples in Fig.\ref{fig:collision_detection} (b). The second stage conducts the same precise collision detection with LLM-A*. Since the AABBs are parallel to the map edge and a path segment does not collide with most of the obstacles, the two-stage method is much simpler and faster than the precise collision detection in LLM-A* ~\cite{zhu2024developing}, which dramatically reduces the total computational overhead for collision detection.

\subsection{Waypoint Generating by Incremental Learning-Based LLM}

The traditional static Few-shot prompts of LLM could not effectively address new environments \cite{song2023llm}, which results in over-fixed planning strategy. To address this issue, an incremental learning mechanism using the robust Few-shot adaptation capacity of LLMs is proposed. Guided by the prompt engineering practices for Qwen models ~\cite{long2025makes} and the Natural Language Processing principle ~\cite{liu2023pre}, the prompts for iLLM-A* include 3 components: Prompt Template, Incremental Learning-Based Few-shot Example Augmentation, and Task Instructions.

\begin{table*}[!t]
\centering
\begin{tabular}{|p{0.95\textwidth}|}
\hline
\textbf{Module I: Static Prompt Template} \\[0.5em]

\textbf{\# Role} \\
You are an expert specializing in computational geometry and path planning. \\[0.5em]

\textbf{\# Goal} \\
Generate an optimal path from start point to goal point based on the given start position, goal position, and obstacle information. \\[0.5em]

\textbf{\# Constraints} \\
Strictly adhere to the following rules: \\
1. \textbf{Obstacle Avoidance}: The path must not contact or intersect any obstacles in any form. \\
2. \textbf{Path Point Count}: The final path must contain at least 5 coordinate points (including start and goal points). \textbf{Interpolation Rules}: If the initially calculated path contains fewer than 5 points, uniformly insert additional points between the longest path segments until the quantity requirement is satisfied. Inserted points should ensure path smoothness and avoid unnecessary sharp angles. \\
3. \textbf{Path Optimality}: Under the premise of satisfying all above constraints, the generated path should approximate the geometrically shortest path. \\[0.5em]

\textbf{\# Input and Output Format} \\
\textbf{Input}: Start point \texttt{start}: \texttt{[x, y]}, Goal point \texttt{goal}: \texttt{[x, y]}, Horizontal barriers \texttt{horizontal\_barriers}: \texttt{[[y, x\_start, x\_end], ...]}, Vertical barriers \texttt{vertical\_barriers}: \texttt{[[x, y\_start, y\_end], ...]} \\
\textbf{Output}: Must strictly follow the JSON array format: \texttt{Generated Path: [[x1, y1], [x2, y2], ..., [xN, yN]]} \\[0.5em]

\textbf{\# Workflow} \\
1. \textbf{Initial Pathfinding}: Based on A* algorithm logic, identify a shortest path that avoids all obstacles. \\
2. \textbf{Verification}: Check the path point count. If fewer than 5 points, add intermediate points according to Core Constraint \\
3. \textbf{Final Validation}: Before output, verify that the generated path completely satisfies all core constraints. \\
\hline
\end{tabular}
\caption{Static Prompt Template for LLM Waypoint Generation.}
\label{tab:static_template}
\end{table*}

\begin{table}[!t]
\centering
\begin{tabular}{|p{0.95\linewidth}|}
\hline
\textbf{Module II: Dynamic Few-shot Examples} \\[0.5em]

\textbf{Input}: \\
start: [94, 321] \\
goal: [706, 668] \\
horizontal\_barriers: [[494, 166, 634], [474, 57, 386]] \\
vertical\_barriers: [[247, 182, 632], [553, 387, 775]] \\
\textbf{Output}: \\
Generated Path: [[94, 321], [217, 211], [341, 275], [464, 387], [588, 421], [650, 544], [706, 668]] \\[0.5em]

[10  In-context demonstrations abbreviated] \\
\hline
\end{tabular}
\caption{Few-shot Examples for  In-context Learning.}
\label{tab:few_shot_demos}
\end{table}

\begin{table}[!t]
\centering
\begin{tabular}{|p{0.95\linewidth}|}
\hline
\textbf{Module III: Task Instructions} \\[0.5em]

Generate intermediate waypoints for the following input. If input data is ambiguous or constraint conditions contain logical conflicts, explicitly identify the problematic areas. Ensure the path generation is completely based on the provided input data. A path that perfectly follows all constraints will be considered a successful response. \\[0.5em]

Start Point: \{start\} \\
goal: \{goal\} \\
horizontal\_barriers: \{horizontal\_barriers\} \\
vertical\_barriers: \{vertical\_barriers\} \\
Generated Path: \\
\hline
\end{tabular}
\caption{Task-Specific Instructions and Input Template.}
\label{tab:task_instructions}
\end{table}

\subsubsection{Prompt Template.}
The template first establishes the LLM's role as a path planning specialist and specifies its fundamental goal to generate an optimal path given the start location, goal locations, and obstacles. Then the template incorporates the key constraints, including obstacle avoidance, minimum required waypoint number (iLLM-A* follows LLM-A* to set this number to $5$), and preferences for geometrically optimal paths. Under these constraints, the generated path is required to approximate the geometrically shortest path. Thereafter, the template standardizes the input/output to follow the JSON format. Finally, the systematic reasoning processes are specified: 1) Using A* to find the collision-free shortest path; 2) Verifying the path point count; 3) Checking that all the constraints are being satisfied. A prompt example is presented in Tab.\ref{tab:static_template}.


\subsubsection{Incremental Learning Based Few-shot Example Augmentation.} A Few-shot example repository contains multiple validated map-waypoints pairs that serve as in-context learning references. These examples undergo incremental updates based on the performance validation outcomes from a set of training maps. 

The incremental learning is to enrich the Few-shot example repository, which enables the LLM to progressively adapt its waypoint generation strategy to diverse environmental characteristics. Given a training map, the LLM generates the waypoints using the current prompt. Then the optimized A* algorithm subsequently searches the path using the waypoints. For clarity, the path is denoted as $\pi_{LLM}$. The optimal path from the start to the goal is also planned using the optimized A* as a baseline, which is denoted $\pi_{base}$. The path length, search time, and memory of $\pi_{LLM}$ are thereafter evaluated:

\begin{equation}
\frac{|Length(\pi_{LLM}) - Length(\pi_{base})|}{|Length(\pi_{base})|} \leq \theta_{length},
\label{eq:lengthth}
\end{equation}
\begin{equation}
\frac{Time(\pi_{LLM})}{Time(\pi_{base})} \leq \theta_{time},
\label{eq:timeth}
\end{equation}
\begin{equation}
\frac{Memory(\pi_{LLM}^{(t)})}{Memory(\pi_{base})} \leq \theta_{memory},
\label{eq:memoryth}
\end{equation}
where $Length(\pi)$ is the path length, $Time(\pi)$ is the search time, and $Memory(\pi)$ is the maximum occupied storage for planning the path $\pi$. If the path $\pi_{LLM}$ satisfies Eq.(\ref{eq:lengthth})-(\ref{eq:memoryth}) simultaneously, the length of the planned path is within $1+\theta_{length}$ of the shortest path, and its search time cost and memory cost are lower than $\theta_{time}$ and $\theta_{memory}$ of the optimized A*, respectively. In iLLM-A*, the 3 thresholds $\theta_{length}$, $\theta_{time}$, and $\theta_{memory}$ are set to $0.1$, which means that the quality of the planned path (and the corresponding waypoints) is extremely high if Eq.(\ref{eq:lengthth})-(\ref{eq:memoryth}) are true. Therefore, the map and the waypoints are incorporated into the Few-shot examples. In this way, the Few-shot example repository is augmented, and the LLM could learn high-quality planning instances and improve its capacity to generate high-quality waypoints for new maps. In this work, we augment the Few-shot example repository with 10 training maps and the LLM-generated waypoints. If more training maps are available, the ones from the oldest training maps are replaced.  Tab.\ref{tab:few_shot_demos} shows a typical Few-shot example repository.

\subsubsection{Task Instructions.} 
This component delivers explicit instructions for current planning queries, integrating specific environmental parameters (start location, goal location, and obstacles) with the template to activate problem-specific reasoning processes. An example of task instruction is presented in Tab.\ref{tab:task_instructions}.

\begin{table}[t]
\centering
\fontsize{8}{10}\selectfont
\renewcommand{\arraystretch}{1.1}
\setlength{\tabcolsep}{6pt}
\begin{tabular}{@{}c|c|cccc@{}}
\hline
\hline
\multicolumn{1}{c|}{\multirow{2}{*}{Metrics}} & \multicolumn{1}{c|}{\multirow{2}{*}{Method}} & \multicolumn{4}{c}{Number of Selected Waypoints} \\
\cline{3-6}
 & & 1 & 2 & 3 & 4 \\
\hline
\multirow{4}{*}{Memory Score ($\uparrow$)} 
& Start & 0.973 & \textbf{0.984} & 0.234 & 0.406 \\
& Uniform & 0.405 & 0.778 & 0.745 & 0.338 \\
& Random & 0.633 & 0.711 & 0.674 & 0.223 \\
& Goal & 0.054 & 0.233 & 0.466 & 0.261 \\
\hline
\multirow{4}{*}{Time Score ($\uparrow$)} 
& Start & 0.972 & \textbf{0.982} & 0.221 & 0.413 \\
& Uniform & 0.420 & 0.780 & 0.743 & 0.374 \\
& Random & 0.354 & 0.716 & 0.662 & 0.238 \\
& Goal & 0.323 & 0.235 & 0.477 & 0.275 \\
\hline
\multirow{4}{*}{Path Length (\%, $\uparrow$)} 
& Start & 107 & 106 & 107 & 107 \\
& Uniform & \textbf{105} & 108 & 108 & 106 \\
& Random & 108 & 107 & 106 & 107 \\
& Goal & 108 & 106 & 107 & 106 \\
\hline
\hline
\end{tabular}
\caption{Waypoint Selection Performance Given Different Methods. \textbf{Bold} Values Indicate The Best Performance.}
\label{tab:waypoint_selection}
\end{table}

\subsection{Appropriate Waypoint Selection}

Even with incremental learning, LLM-generated waypoints may contain redundant or deviating ones that compromise the efficiency of the A* algorithm. Inspired by the work ~\cite{karaman2011anytime}, we develop an empirically validated method to select the appropriate subset of waypoints.

\subsubsection{Empirical Study.} Similarly to LLM-A*, we make LLM generate at least $5$ waypoints, the experiments demonstrate that excessive waypoints introduce extra computational overhead. 
We compare 4 waypoint subset selection methods: Uniform Selection (uniformly choosing some of the waypoints, abbr. as Uniform), Start-Prioritized Selection (prioritizing the waypoints closer to the start, abbr. as Start), Goal-Prioritized Selection (prioritizing the waypoints closer to the goal, abbr. as Goal), and Random Selection (randomly choosing some of the waypoints, abbr. as Random). With these methods, we choose 1-4 waypoints and compare their performance in Tab.\ref{tab:waypoint_selection}, with each experimental group repeated for 30 runs. The details of the score calculation are described in the Appendix.


\subsubsection{Empirical Results Analysis.}

Tab.\ref{tab:waypoint_selection} shows the Start method achieves the highest scores of 0.973/0.984 for memory and 0.972/0.982 for time efficiency using 1-2 waypoints, while maintaining path path within only $6\%-7\%$ to the optimal path length. The performance of all the selection methods significantly degrades given $>2$ waypoints.

\subsubsection{Waypoint Selection.}
Inspired by the empirical results above, we select the appropriate waypoints as follows: If the number of the LLM-generated waypoints is not larger than 2, all waypoints are utilized for path planning; if the number exceeds 2, the first two waypoints closest to the start location are selected.



\begin{table}[t]
\centering
\fontsize{9}{11}\selectfont
\renewcommand{\arraystretch}{1.4}
\setlength{\tabcolsep}{1pt}
\begin{tabular}{c|c|cccccc}
\hline\hline
\multirow{2}{*}{\centering Metrics}
& \multicolumn{1}{c|}{\multirow{2}{*}{\centering Method}}
& \multicolumn{6}{c}{Map Size ($N$)} \\
\cline{3-8}
& & 200 & 250 & 300 & 350 & 400 & 450 \\
\hline
\multirow{5}{*}[-0.5mm]{\rotatebox{90}{\makebox[1.3cm][c]{Path Length (\%)}}}
& \centering iLLM-A* & 102.83 & 109.46 & 105.29 & 106.17 & 107.00 & 109.40 \\
& \centering w/o LLM & 100.00 & 100.00 & 100.00 & 100.00 & 100.00 & 100.00 \\
& \centering w/o IncL & 102.83 & 113.86 & 106.57 & 107.40 & 114.80 & 115.04 \\
& \centering w/o WDS & 109.03 & 115.10 & 104.50 & 105.00 & 108.25 & 108.52 \\
& \centering w/o Opt-A* & 108.74 & 106.30 & 105.66 & 107.35 & 117.20 & --- \\
\hline
\multirow{5}{*}[-0.5mm]{\rotatebox{90}{\makebox[1.3cm][c]{Search Time (s)}}}
& \centering iLLM-A* & \textbf{0.11} & \textbf{0.21} & \textbf{0.33} & \textbf{0.41} & \textbf{0.39} & \textbf{0.79} \\
& \centering w/o LLM & 0.50 & 0.66 & 1.62 & 2.83 & 3.75 & 9.05 \\
& \centering w/o IncL & 0.12 & 0.21 & 0.33 & 0.47 & 0.59 & 0.96 \\
& \centering w/o WDS & 0.13 & 0.39 & 1.50 & 3.04 & 2.49 & 4.33 \\
& \centering w/o Opt-A* & 27.04 & 54.17 & 173.54 & 278.96 & 381.96 & --- \\
\hline
\multirow{5}{*}[-0.5mm]{\rotatebox{90}{\makebox[1.3cm][c]{Memory (MB)}}}
& \centering iLLM-A* & \textbf{1.02} & \textbf{1.57} & \textbf{2.14} & \textbf{2.23} & \textbf{2.54} & \textbf{3.62} \\
& \centering w/o LLM & 3.35 & 3.61 & 8.69 & 13.37 & 14.10 & 28.67 \\
& \centering w/o IncL & 1.21 & 1.74 & 2.41 & 2.73 & 2.79 & 3.98 \\
& \centering w/o WDS & 1.34 & 1.93 & 5.23 & 6.52 & 5.47 & 6.13 \\
& \centering w/o Opt-A* & 1.06 & 1.63 & 2.14 & 2.28 & 2.62 & --- \\
\hline\hline
\end{tabular}
\caption{Ablation Study on Large-scale Maps. "---" indicates Search Time$>10 min$.}
\label{tab:ablation_study}
\end{table}

\begin{table*}[h]
\centering
\renewcommand{\arraystretch}{1.1}
\setlength{\tabcolsep}{4pt}
\fontsize{10}{12}\selectfont
\resizebox{1\textwidth}{!}{
\begin{tabular}{c|cccc|cccc|cccc}
\hline
\hline
\multirow{3}{*}{\centering Map Size ($N$)} & \multicolumn{12}{c}{Performance Metrics} \\
\cline{2-13}
& \multicolumn{4}{c|}{Search Time (s) $\downarrow$} & \multicolumn{4}{c|}{Memory (MB) $\downarrow$} & \multicolumn{4}{c}{Path Length (\%) $\downarrow$} \\
\cline{2-13}
& A* & Opt-A* & LLM-A* & iLLM-A* & A* & Opt-A* & LLM-A* & iLLM-A* & A* & Opt-A* & LLM-A* & iLLM-A* \\
\hline
50 & 0.06 & 0.002 & 0.08 & \textbf{0.001} & 0.067 & 0.075 & 0.061 & \textbf{0.059} & 100.00 & 100.00 & 110.47 & 100.01 \\
100 & 1.11 & 0.07 & 0.91 & \textbf{0.04} & 1.22 & 1.84 & 1.13 & \textbf{0.64} & 100.00 & 100.00 & 101.69 & 101.03 \\
150 & 27.24 & 0.20 & 16.52 & \textbf{0.11} & 1.79 & 2.22 & 1.96 & \textbf{0.98} & 100.00 & 100.00 & 106.86 & 108.5 \\
200 & 89.71 & 0.50 & 34.65 & \textbf{0.11} & 3.32 & 3.35 & 2.43 & \textbf{1.02} & 100.00 & 100.00 & 112.3 & 103.26 \\
250 & 245.46 & 0.66 & 229.15 & \textbf{0.21} & 3.42 & 3.61 & 2.75 & \textbf{1.57} & 100.00 & 100.00 & 108.22 & 108.63 \\
300 & --- & 1.62 & 438.86 & \textbf{0.33} & --- & 8.69 & 5.17 & \textbf{2.14} & --- & 100.00 & 104.04 & 105.93 \\
350 & --- & 2.83 & --- & \textbf{0.41} & --- & 13.37 & --- & \textbf{2.23} & --- & 100.00 & --- & 106.17 \\
400 & --- & 3.75 & --- & \textbf{0.39} & --- & 14.10 & --- & \textbf{2.54} & --- & 100.00 & --- & 107.37 \\
450 & --- & 9.05 & --- & \textbf{0.79} & --- & 28.67 & --- & \textbf{3.62} & --- & 100.00 & --- & 108.00 \\
\hline
\hline
\end{tabular}
}
\caption{Performance Given Different Map Sizes. "---" Indicates Search Time $>10min$.}
\label{tab:main_results}
\end{table*}

\section{Evaluation}
\label{sec:eva}
\subsection{Experiment Settings}
\noindent\textbf{Grid Map Settings.} The evaluations are conducted on different kinds of square 2D grid maps. Specifically, we leverage the LLM to generate a series of maps whose edge length $N$ are \{$50, 100, 150, 200, 250, 300, 350, 400, 450$\}. For the smallest map ($N=50$), we set the obstacle number to 3. For the larger ones, the obstacle number increases linearly with the map area. For example, the obstacle number of the map with $N=450$ is $3\times (450/50)^2=243$. The width of the obstacle equals 1 grid, and the length of the obstacle is randomly distributed between $10$ and $50$ grids. The obstacles are randomly distributed in the map, which is randomly parallel to the vertical or horizontal edge of the map. To evaluate the robustness of iLLM-A* in diverse maps, we insert two new kinds of giant obstacles in the maps, which will be described in details in Sec.\ref{sec:obstacle}. The numerical results are the average of 30 runs.

\noindent\textbf{Baselines.} The evaluation includes 3 baselines:
\begin{itemize}
\item\textbf{ LLM-A*}: LLM-A* \cite{meng2024llm} represents the SOTA method wherein the LLM generates intermediate waypoints to guide A* to search.
\item \textbf{ A*}: The original A* algorithm employed in LLM-A*.
\item\textbf{Opt-A*}: The optimized A* in iLLM-A*.

\end{itemize}
\noindent\textbf{Evaluation Metrics.} We compare iLLM-A* on 3 metrics.
\begin{itemize}

\item \textbf{ Path Length (\%).} Since A* and Opt-A* could find the shortest path, we normalized the length of the shortest path as $100\%$. The planned path by LLM-A* and iLLM-A* should not be shorter than $100\%$.
\item \textbf{Search Time (s/seconds).} This is the time consumed by an algorithm for planning the path.
\item \textbf{Memory (MB).} This is the maximum storage needed for an algorithm when planning the path.
\end{itemize}



\noindent\textbf{Environment.} We implement iLLM-A* and the baselines in Python on a server equipped with Intel Xeon Silver 4216 CPU, NVIDIA V100 GPU, and 128 GB RAM running Ubuntu with CUDA 550. Qwen2.5-32B is used as the LLM.  

\subsection{Ablation Study}
Tab.\ref{tab:ablation_study} shows the Path Length, Search Time, and Memory of the variants on different map sizes. iLLM-A* outperforms all variants. Without LLM, the Search Time is obviously longer than that of iLLM-A* (e.g. $9.05s$ vs. $0.79s$ for the map with $N=450$). Without IncL, the average Path Length is $3.2\%$ longer than that of iLLM-A*. Meanwhile, the maximum Path Length in the variant without Incl is $115.05\%$, which is much worse than that of iLLM-A*. Without WDS, both the performance on Search Time and Memory are obviously worse than that of iLLM-A*. Without Opt-A*, the Search Time is $270\times$ to $1000\times$ longer than that of iLLM-A*. The ablation study shows the effectiveness of each designed mechanism in iLLM-A*.

\subsection{Overall Performance Comparison}

\begin{figure}[t]
    \centering
    \includegraphics[width=0.9\columnwidth]{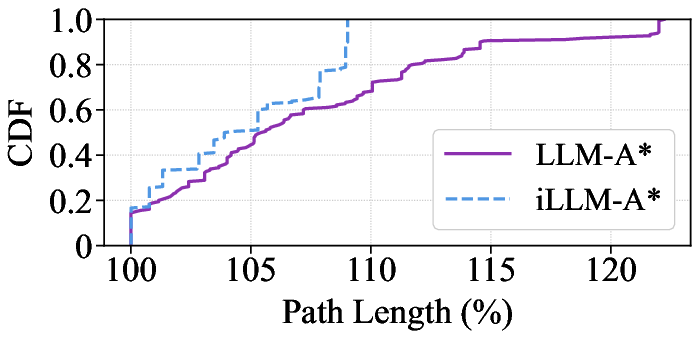}
    \caption{CDF of Path Length For LLM-A* and iLLM-A*.}
    \label{fig:cdf_comparison}
\end{figure}

\begin{table}[t]
\centering
\fontsize{8}{10}\selectfont
\renewcommand{\arraystretch}{1.1}
\setlength{\tabcolsep}{9pt}
\begin{tabular}{@{}c|cccccc@{}}
\hline
\hline
\multicolumn{1}{c|}{\multirow{2}{*}{Method}} & \multicolumn{6}{c}{Map Size (N)} \\
\cline{2-7}
 & 50 & 100 & 150 & 200 & 250 & 300 \\
\hline
iLLM-A* & \textbf{0.00} & \textbf{0.27} & \textbf{0.64} & \textbf{4.82} & \textbf{0.46} & \textbf{1.20} \\
LLM-A* & 0.63 & 1.67 & 3.06 & 7.97 & 5.56 & 3.46 \\
\hline
\hline
\end{tabular}
\caption{Path Length Std on Different Map Sizes (\%)$\downarrow$.}
\label{tab:robustness_comparison}
\end{table}

\begin{figure}[t]
    \centering
    \includegraphics[width=0.9\columnwidth]{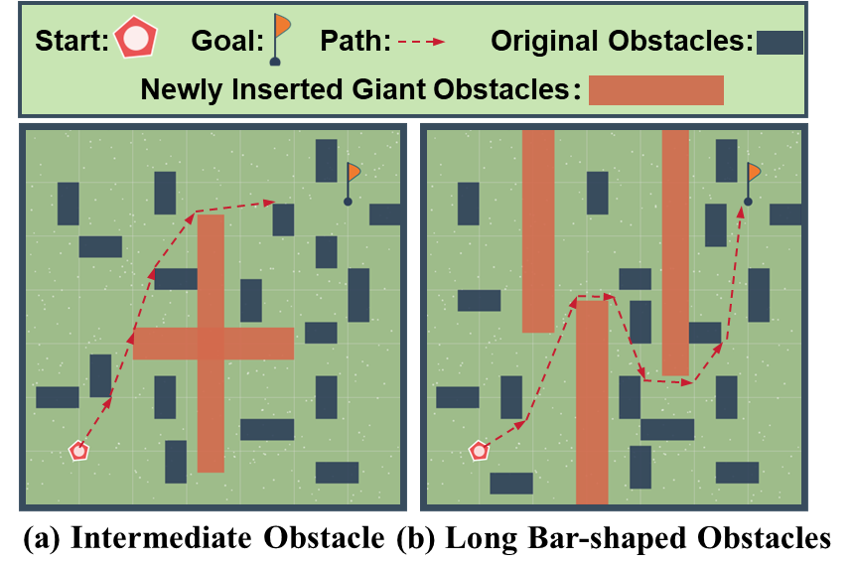}
    \caption{Basic Concept of the Two New Kinds of Obstacles.}
    \label{fig:challenge_scenarios}
\end{figure}

\begin{table*}[t]
\centering
\renewcommand{\arraystretch}{1.1}
\setlength{\tabcolsep}{4pt}
\fontsize{10}{12}\selectfont
\resizebox{1\textwidth}{!}{
\begin{tabular}{c|cccc|cccc|cccc}
\hline
\hline
\multirow{3}{*}{\centering Map Size ($N$)} & \multicolumn{12}{c}{Performance Metrics} \\
\cline{2-13}
& \multicolumn{4}{c|}{Search Time (s) $\downarrow$} & \multicolumn{4}{c|}{Memory (MB) $\downarrow$} & \multicolumn{4}{c}{Path Length (\%) $\downarrow$} \\
\cline{2-13}
& A* & Opt-A* & LLM-A* & iLLM-A* & A* & Opt-A* & LLM-A* & iLLM-A* & A* & Opt-A* & LLM-A* & iLLM-A* \\
\hline
50 & 0.51 & 0.01 & 0.04 & \textbf{0.002} & 0.078 & 0.15 & 0.14 & \textbf{0.07} & 100.00 & 100.00 & 102.20 & 101.29 \\
100 & 4.86 & 0.05 & 3.05 & \textbf{0.01} & 0.75 & 0.85 & 0.44 & \textbf{0.30} & 100.00 & 100.00 & 107.22 & 103.75 \\
150 & 29.35 & 0.18 & 19.30 & \textbf{0.10} & 1.78 & 2.23 & 2.74 & \textbf{1.11} & 100.00 & 100.00 & 107.33 & 105.83 \\
200 & 152.01 & 0.75 & 145.92 & \textbf{0.12} & 3.81 & 3.15 & 3.18 & \textbf{1.34} & 100.00 & 100.00 & 105.85 & 104.80 \\
250 & 170.48 & 0.585 & 375.92 & \textbf{0.16} & 3.94 & 3.67 & 3.87 & \textbf{1.56} & 100.00 & 100.00 & 104.74 & 109.28 \\
300 & --- & 1.40 & --- & \textbf{0.37} & --- & 8.16 & --- & \textbf{2.45} & --- & 100.00 & --- & 104.82 \\
350 & --- & 2.637 & --- & \textbf{0.842} & --- & 13.35 & --- & \textbf{3.23} & --- & 100.00 & --- & 107.91 \\
400 & --- & 5.477 & --- & \textbf{0.605} & --- & 17.36 & --- & \textbf{3.28} & --- & 100.00 & --- & 102.83 \\
450 & --- & 8.733 & --- & \textbf{0.800} & --- & 28.69 & --- & \textbf{3.37} & --- & 100.00 & --- & 109.42 \\
\hline
\hline
\end{tabular}
}
\caption{Performance For The Map With One Intermediate Cross-shaped Obstacle.}
\label{tab:scenario1}
\end{table*}

\begin{table*}[t]
\centering
\renewcommand{\arraystretch}{1.1}
\setlength{\tabcolsep}{4pt}
\fontsize{10}{12}\selectfont
\resizebox{1\textwidth}{!}{
\begin{tabular}{c|cccc|cccc|cccc}
\hline
\hline
\multirow{3}{*}{\centering Map Size ($N$)} & \multicolumn{12}{c}{Performance Metrics} \\
\cline{2-13}
& \multicolumn{4}{c|}{Search Time (s) $\downarrow$} & \multicolumn{4}{c|}{Memory (MB) $\downarrow$} & \multicolumn{4}{c}{Path Length (\%) $\downarrow$} \\
\cline{2-13}
& A* & Opt-A* & LLM-A* & iLLM-A* & A* & Opt-A* & LLM-A* & iLLM-A* & A* & Opt-A* & LLM-A* & iLLM-A* \\
\hline
50 & 0.77 & 0.012 & 0.19 & \textbf{0.003} & 0.33 & 0.39 & 0.24 & \textbf{0.15} & 100.00 & 100.00 & 108.78 & \textbf{100.00} \\
100 & 6.44 & 0.07 & 5.34 & \textbf{0.027} & 0.84 & 0.94 & 0.87 & \textbf{0.56} & 100.00 & 100.00 & 100.65 & \textbf{100.83} \\
150 & 31.55 & 0.22 & 25.31 & \textbf{0.11} & 1.79 & 2.22 & 1.78 & \textbf{1.05} & 100.00 & 100.00 & 110.06 & \textbf{107.67} \\
200 & 134.49 & 0.61 & 125.43 & \textbf{0.13} & 3.51 & 3.89 & 3.28 & \textbf{1.12} & 100.00 & 100.00 & 104.95 & \textbf{109.11} \\
250 & 217.65 & 0.585 & 436.23 & \textbf{0.27} & 3.75 & 4.08 & 5.14 & \textbf{2.11} & 100.00 & 100.00 & 106.86 & \textbf{108.41} \\
300 & 523.43 & 1.91 & --- & \textbf{0.27} & 7.25 & 9.08 & --- & \textbf{2.08} & 100.00 & 100.00 & --- & \textbf{103.20} \\
350 & --- & 2.96 & --- & \textbf{0.50} & --- & 13.40 & --- & \textbf{2.38} & --- & 100.00 & --- & \textbf{105.22} \\
400 & --- & 5.477 & --- & \textbf{0.58} & --- & 17.36 & --- & \textbf{2.64} & --- & 100.00 & --- & \textbf{107.75} \\
450 & --- & 8.733 & --- & \textbf{0.76} & --- & 28.69 & --- & \textbf{2.89} & --- & 100.00 & --- & \textbf{108.54} \\
\hline
\hline
\end{tabular}
}
\caption{Performance For The Maps with Long Bar-shaped Obstacles.}
\label{tab:scenario2}
\end{table*}

Tab.\ref{tab:main_results} demonstrates the comprehensive performance given different map sizes. iLLM-A* significantly outperforms all baselines in Search Time and Memory while maintaining acceptable path length. 

\noindent\textbf{Search Time Comparison.} 
iLLM-A* achieves substantial improvements in runtime performance with order-of-magnitude speedups. Specifically, for maps with $N=250$ and $N=300$, iLLM-A* achieves  $1091\times$ ($0.21s$ vs. $229.15s$) and $1330\times$ ($0.33s$ vs. $438.86s$) faster than LLM-A*, respectively. Besides, iLLM-A* achieves $11.5\times$ ($0.79s$vs.$9.05s$) speedup compared with Opt-A* given the map with $N=450$. 


\noindent\textbf{Memory Comparison.} The occupied memory of iLLM-A* is only 3.62 MB vs. 28.67 MB of Opt-A* on the map with $N=450$, representing $87.4\%$ of memory reduction. iLLM-A* saves $58.6\%$ of the memory compared to LLM-A* on the map with $N=300$. This efficiency stems from our waypoint selection method that effectively reduces the exploration space of A*.

\noindent\textbf{Path Length Comparison.} On average, the mean path length of LLM-A* and iLLM-A* is {$107.94\%$ and $104.39\%$} given $N=50,100,150,200,250,300$. Namely, iLLM-A* reduces the gap to the optimal path length by $\frac{7.94-4.39}{7.94}=44.7\%$ compared with LLM-A*, which means the path length of iLLM-A* is much closer to the optimal path. This is because the incremental learning and waypoint selection mechanisms generates higher-quality waypoints.

\subsection{Stability Analysis}
Given the maps with $N=50,100,150,200,250,300$, the Cumulative Distribution Function (CDF) of the Path Length of LLM-A* and iLLM-A* is shown in Fig.\ref{fig:cdf_comparison}. All Path Lengths of iLLM-A* are shorter than $110\%$, while about $30\%$ of the Path Lengths of LLM-A* are longer than $110\%$. To further verify the stability of iLLM-A*, Tab.\ref{tab:robustness_comparison} illustrates the Path Length std in different sizes of maps. Clearly, the standard deviation of the Path Length of iLLM-A* is much smaller than that of LLM-A* for all map sizes, which shows that the length of the paths planned by iLLM-A* is more consistent and stable to the optimal path.

\subsection{Scalability Analysis}\label{sec:obstacle}

To show the robustness of iLLM-A*, we insert two new kinds of giant obstacles in the maps, as illustrated in Fig.\ref{fig:challenge_scenarios}. Specifically, one giant cross-shaped obstacle lies in the intermediate region between the start of the goal, which forces the feasible paths to detour far from the straight line connecting the start and goal. The other kind of obstacle includes 3 long bar-shaped obstacles parallel to an edge, which force the feasible paths to detour multiple times. The length of the obstacles is randomly distributed within $50\%$-$60\%$ of the edge length.  

Tab.\ref{tab:scenario1} shows that iLLM-A* achieves near-linear scalability for the map with one intermediate giant obstacle, with search times scaling linearly with the map size, ranging from $0.002s$ to $0.800s$ across all map sizes. In contrast, LLM-A* cannot complete the planning tasks on the map with $N\geq 300$. Specifically, given $N=250$, the Search Time of LLM-A* is $2349.5\times$ longer than that of iLLM-A* ($375.92s$ vs. $0.16s$). The Path Length consistently remains within $100\%$ and $109. 42\%$. The Memory of iLLM-A* is also only about half of that of LLM-A*. The results for maps with long bar-shaped obstacles are similar and are illustrated in Tab.\ref{tab:scenario2}. For maps with $N=200, 250$, the search time of LLM-A* is about $1000\times$ longer than that of iLLM-A*. The results are consistent with those in Tab.\ref{tab:main_results}, which shows the robustness of iLLM-A* in diverse maps, even with unseen giant obstacles in the Few-shot examples.

\section{Conclusion}
In this paper, we present an innovative LLM-enhanced algorithm for path planning in large-scale grid maps. We first use both test and theoretical analysis to reveal the limitations of the SOTA LLM-enhanced algorithm. Then we propose the innovative design consisting of 3 core mechanisms to address these limitations: Optimization of A*, Waypoint Generating By Incremental Learning LLM, and Appropriate Waypoint Selection. Finally, a comprehensive evaluation on various grid maps shows that, compared with SOTA method, iLLM-A* \textbf{1) achieves more than $1000\times$ speedup on average, 2) {saves up to $58.6\%$ of the memory cost}, 3) achieves both obviously shorter path length and lower path length standard deviation.}


\bibliography{aaai25}

\clearpage

\appendix            
\section{Appendix: Metric Computation and Normalization}
\label{appendix:A2}

The experimental evaluation implements a two-stage metric computation process involving raw data aggregation and normalization-based scoring procedures.

\subsubsection{Raw Metric Aggregation}: The system calculates the arithmetic mean of the three performance metrics across 30 trials for each strategy-count-map combination.
\subsubsection{Normalization for Cost-Type Metrics}: Search time and memory constitute cost-type indicators where lower values demonstrate superior performance. The normalization process applies the following mathematical transformation for each map scale:

\begin{equation}
Score_{normalized} = \frac{x_{max} - x_{current}}{x_{max} - x_{min}}
\end{equation}

where $x_{current}$ represents the mean performance value for the specific strategy-count combination, $x_{max}$ denotes the maximum value across all strategies for the given map scale, and $x_{min}$ represents the corresponding minimum value. This transformation maps performance values to a [0,1] scale where values approaching 1.0 indicate optimal performance characteristics.

\subsubsection{Final Score Calculation}: The system computes the final score for each cost-type metric by averaging the normalized scores across all map scales, ensuring balanced representation across different complexity levels.

\subsubsection{Path Length Evaluation}: The path length assessment calculates the Path Length as a percentage of the optimal A* solution for each individual trial, subsequently computing the mean Path Length across all trials and map scales for each strategy-count combination. Values approaching 100\% indicate superior path quality performance.

\clearpage

\end{document}